\pdfoutput=1

\documentclass[11pt]{article}

\usepackage[]{ACL2023}

\usepackage{times}
\usepackage{latexsym}
\usepackage[T1]{fontenc}
\usepackage[utf8]{inputenc}
\usepackage{microtype}
\usepackage{inconsolata}

\usepackage{multirow}
\usepackage{booktabs}
\usepackage{hyperref}
\usepackage{amsmath}
\usepackage{amssymb}
\usepackage{framed}
\usepackage{graphicx}
\usepackage{array}

\title{Intent Induction from Conversations for Task-Oriented Dialogue Track at DSTC 11}

\author{
    \textbf{\large James Gung, Raphael Shu, Emily Moeng, Wesley Rose, Salvatore Romeo}\\
    \textbf{\large Yassine Benajiba, Arshit Gupta, Saab Mansour \and Yi Zhang}\\
    AWS AI Labs \\
    \texttt{\{gungj,zhongzhu,emimoeng,rosewes,romeosr,arshig,benajiy,saabm,yizhngn\}@amazon.com}\\
}

\newcommand{\ours}{\textsc{NatCS}}

\newcommand{\oursbankingdstc}{Banking}
\newcommand{\oursfinancedstc}{Finance}
\newcommand{\oursinsuranceselfdstc}{Insurance}

\newcommand{\sgd}{SGD}

\newcommand{\multiwoz}{MWOZ}

\newcommand{\clincoos}{CLINC150}
\newcommand{\polyaibank}{BANK77}

\newcommand{\da}[1]{\textit{#1}}
\newcommand{\plusminus}[1]{$ {\scriptscriptstyle\pm{#1}}$}
\newcommand{\std}[2]{#1\plusminus{#2}}

\newcommand\InformIntent{\da{InformIntent}}

\newcommand\fone{$\text{F}_1$}
\newcommand\mrr{$\text{MRR}_{avg}$}

\begin{document}
\maketitle
\begin{abstract}

With increasing demand for and adoption of virtual assistants, recent work has investigated ways to accelerate bot schema design through the automatic induction of intents or the induction of slots and dialogue states.
However, a lack of dedicated benchmarks and standardized evaluation has made progress difficult to track and comparisons between systems difficult to make.
This challenge track\footnote{\url{https://github.com/amazon-science/dstc11-track2-intent-induction}}, held as part of the Eleventh Dialog Systems Technology Challenge, introduces a benchmark that aims to evaluate methods for the automatic induction of customer intents in a realistic setting of customer service interactions between human agents and customers.
We propose two subtasks for progressively tackling the automatic induction of intents and corresponding evaluation methodologies.
We then present three datasets suitable for evaluating the tasks and propose simple baselines.
Finally, we summarize the submissions and results of the challenge track, for which we received submissions from 34 teams.
 
\end{abstract}

\section{Introduction}
\label{sec:introduction}
Task-oriented dialogue systems used for handling high-volume customer service requests have seen growing adoption in recent years.
While numerous platforms for building such dialogue systems exist, they still typically require significant time and expert knowledge to achieve good results.
In particular, the process of creating non-overlapping intents and corresponding example utterances typically requires domain expertise and/or laborious analysis of a large volume of conversation transcripts.
Intent induction, automatically deriving intents and corresponding utterances from conversation transcripts or customer queries, has the potential to significantly reduce the time and effort required to build a task-oriented dialogue system from the ground up.

Recent work in intent mining has often cast intent induction as a problem similar to that of short text clustering, in which unlabeled customer queries and requests are assigned labels and the effectiveness of a system is evaluated by comparing cluster assignments to labels from a ground truth intent schema~\cite{hakkani2015clustering,haponchyk-etal-2018-supervised,perkins-yang-2019-dialog,chatterjee-sengupta-2020-intent}.
Due to the lack of real world datasets capturing human-human customer support conversations, researchers have repurposed existing intent classification datasets or labeled small subsets of publicly available customer service interactions to evaluate their systems.

Despite a growing body of work, there is a lack of common evaluative settings and standardized, dedicated benchmarks for intent induction, making progress difficult to track.
Existing intent classification datasets do not capture the complexity of mining intents from real customer service interactions, which typically have highly skewed distributions of intents that are embedded in noisy conversations.
While dedicated, representative benchmarks have been instrumental in driving and measuring progress in natural language processing tasks~\cite{wang2019glue,ruder2021benchmarking}, there is a clear gap in this regard for intent induction.

To encourage further research and to provide a shared benchmark in the realistic setting of spoken customer service interactions, this challenge track introduces a dataset containing conversations spanning three domains: insurance, personal banking, and finance.
The track explores an alternative framing of the intent induction task through the use of two related subtasks: Task 1) \emph{intent clustering} and Task 2) \emph{open intent induction}.
For the intent clustering task, we use classic clustering metrics for evaluation.
For open intent induction, which seeks to provide a more realistic and noisy setting in which intents are embedded in conversations, we evaluate systems by examining the predictions of an intent classifier trained on induced intent schemas.
This setting aims to bring us closer to assessing the impact of automatic induced intents on a final dialogue system.

A total of 34 teams participated in the track, with 19 teams also participating in Task 2.
In this paper, we describe the tasks, evaluation methods, datasets, and baselines. 
We then describe the submissions from the participating teams and summarize the results and findings from the track.

\section{Related Work}
\label{sec:relatedwork}
The majority of existing work in mining intents has focused on clustering-based methods.
\citet{perkins-yang-2019-dialog} proposes a multi-view clustering approach for learning clustering representations by predicting cluster assignments of an alternative view of each input, such as prompts.
\citet{chatterjee-sengupta-2020-intent} investigates variants of DBSCAN and propose an approach that iteratively breaks down the ``noise'' cluster from DBSCAN to address varying densities. 
Others have leveraged intermediate structured prediction tasks (such as dependency parsing or abstract meaning representations) to aid in the induction of intents~\cite{hakkani2015clustering,vedula2020open, zeng2021automatic, liu2021open}.

Prior work in intent mining has largely evaluated systems by re-purposing existing datasets commonly used for evaluating intent classification systems.
Such datasets, including BANK77~\cite{casanueva-etal-2020-efficient}, CLINC150~\cite{larson-etal-2019-evaluation}, SNIPS~\cite{coucke2018snips}, ATIS~\cite{hemphill-etal-1990-atis}, or StackOverflow~\cite{predict-closed-questions-on-stack-overflow,xu-etal-2015-short}, do not include full dialogues.
Task-oriented dialogue datasets like MultiWOZ~\cite{budzianowski-etal-2018-multiwoz}, MultiDoGO~\cite{peskov-etal-2019-multi} and SGD~\cite{rastogi2020towards} span multiple domains, but individual domains contain few user intents, and conversations are not designed to be representative of real human-to-human customer service interactions.
\citet{perkins-yang-2019-dialog} re-purpose two-turn customer support exchanges from Twitter, but only annotate a small subset of dialogues across 14 intents with broad semantics.

\citet{hakkani2015clustering} evaluate the classification performance of induced intents after undergoing manual mappings to reference intents.
More recent work has used clustering metrics commonly used for evaluation of short text clustering in which the number of clusters is provided, such as clustering ACC, NMI, and ARI~\cite{peskov-etal-2019-multi, zhang2021discovering,chatterjee-sengupta-2020-intent,kumar-etal-2022-intent}.
Recent work has also investigated the intent discovery problem in which systems must discover novel intents based on a set of pre-existing 
intents~\cite{lin2020discovering,zhang2021deep,zhang2021discovering,shen-etal-2021-semi,kumar-etal-2022-intent}.
In contrast to this work, this benchmark focuses on the fully unsupervised setting in which no pre-existing intents are defined and the number of reference intents are not provided in advance.
\section{Tasks and Evaluation}
\label{sec:tasks}
\begin{table*}[t]
\small
\centering
\begin{tabular}{rp{0.41\linewidth}p{0.41\linewidth}}
\toprule
 & \textbf{Task 1 - Intent Clustering}  & \textbf{Task 2 - Open Intent Induction}\\
\midrule
\textbf{Goal} & Assign cluster labels to a list of turns & Induce intents and training examples for an intent classifier from conversations \\
\midrule
\textbf{Input} & (1) Conversation transcripts, (2) clustering turns  & (1) Conversation transcripts, (2) Automatic \InformIntent{} turn predictions \\
\midrule
\textbf{Output} &
Intent cluster labels assigned to each clustering turn &
Example utterances labeled with induced intents
\\
\bottomrule
\end{tabular}
\caption{Summary of the benchmark tasks proposed as part of the DSTC 11 challenge on intent induction from conversations: Intent Clustering and Open Intent Induction.}
\label{table:tasks}
\end{table*}

The track consists of two subtasks providing alternative ways of framing and evaluating intent induction.
This section describes the motivation for these tasks and metrics used for evaluating them.

\subsection{Task 1 - Intent Clustering}
Task 1 is a conversational intent clustering task.
In this task, a set of conversation transcripts are provided as inputs, with each turn pre-labeled with a speaker role (i.e. \textit{agent} or \textit{customer}).
Turns within these transcripts that contain intents are tagged for use in the task.
Participants must assign each of these intentful turns to a cluster.
Submissions are evaluated by comparing the resulting cluster assignment with intent labels from a reference intent schema.
The number of reference intents is not provided for the task, though lower and upper limits are given (5 and 50 respectively).

\subsection{Task 2 - Open Intent Induction}
The clustering evaluation in Task 1 has several shortcomings.
In a real world setting, the exact turns in transcripts containing intents are unknown.
The goal of intent induction is to extract a set of distinctly meaningful intents, rather than assigning each turn a cluster label.
The quality of a set of induced intents is likely to be judged based on the coverage and accuracy of a resulting intent classifier rather than coverage on input turns.
From the perspective of chat bot development, a smaller but cleaner training set can thus be preferable to a larger but noisier and less manageable set of utterances.

To account for this, the goal of Task 2 is to instead generate a \textit{training set} for an intent classifier given a set of unlabeled conversations.
The training set consists of utterances derived from the conversations along with corresponding intent labels\footnote{In this track, the intent labels are treated as unique IDs and are not evaluated for linguistic meaning.}.
Note that no explicit correspondence between utterances in the training set and the original conversations is required, so techniques such as data augmentation or paraphrasing are allowed.
Unlike Task 1, to better reflect challenges of noisy data in a real world setting, we do not explicitly provide labels to tell whether a turn is intentful.
However, to simplify system development for the task, automatic dialogue act classifier predictions for \da{InformIntent} are provided to participants.

Evaluation for the task is conducted using the predictions of an intent classifier trained with the induced intents.
After training the classifier, predictions are made on a separate test set of utterances labeled with a reference intent schema.
Finally, similarly to Task 1, the quality of the induced set of intents is evaluated by comparing the predicted assignment of labels with the reference intent assignment.
In contrast to \citet{hakkani2015clustering}, this evaluation approach is fully automatic and does not require manual intent mappings.

\subsection{Metrics}
\label{sec:evaluation}
In a realistic setting, the number of unique intents that are present in a dataset will not be known in advance. 
Inducing an excess of fine-grained intents may result in purer, more coherent predictions, but will then require additional manual effort and human expertise to merge intents that are duplicates of one another (semantically equivalent).
On the other hand, predicting intents with too broad or coarse-grained semantics is also undesirable.
We therefore select metrics that balance this trade-off and encourage approaches capable of matching the granularity of the reference intent schema in the test data.

Metrics for both Task 1 and Task 2 are computed by comparing the assignment of induced intents, $C$, with assignments to a single reference intent schema, $L$.
For Task 1, cluster assignments are compared on turns in input conversations.
For Task 2, predicted induced intents are compared on a set of test utterances collected independently from input conversations.

Clustering accuracy (ACC) is a commonly used metric for short text clustering that penalizes solutions for producing either too many or too few clusters~\cite{huang2014deep}.
ACC is defined as
\begin{align*}
    \text{ACC} = \frac{\sum^N_{i=1}\delta(map({\hat{y}}_i)={y}_i)}{N} ,
\end{align*}
where $\delta(\cdot)$ is an indicator function that outputs 1 when the argument is true or 0 when false, ${\hat{y}}_i \in C$ and ${y}_i \in L$ are the predicted and ground truth labels for the $i$th input respectively, and $N$ is the total number of turns/inputs.
The $map(\cdot)$ function assigns the cluster label to the optimal label ${y}_i$ as computed by the Hungarian algorithm~\cite{kuhn1955hungarian}.
If too few intents are predicted, some reference intents will not receive assignments, whereas an excess of induced intents will lead to unassigned induced intents.

Clustering \fone{}~\cite{artiles-etal-2007-semeval,haponchyk-etal-2018-supervised} also captures this trade-off by combining clustering \textit{precision} (purity) and clustering \textit{recall} (inverse purity, in which each reference intent is assigned to the most frequently co-occurring induced intent):
\begin{align*}
\text{precision (P)} &= \frac{\sum^{\lvert C \rvert}_{k=1} \text{max}^{\lvert L \rvert}_{j=1} \lvert l_j \cap c_k \rvert}{N} , \\
\text{recall (R)} &= \frac{\sum^{\lvert L \rvert}_{k=1} \text{max}^{\lvert C \rvert}_{j=1} \lvert l_k \cap c_j  \rvert}{N},
\end{align*}
where $\lvert l_k \cap c_j  \rvert$ indicates the size of the set containing inputs assigned to both reference cluster $l_k$ and predicted cluster $c_j$.
Clustering $\text{F}_1$ is then computed as the harmonic mean of the two measures.
Solutions with too few intents, or broad intents that are split between multiple different reference intents, will have lower precision/purity.
Solutions with an excess of granular intents mapping to the same reference intent will have lower recall, as each reference intent can only be assigned to a single induced intent.

Finally, we also report NMI (normalized mutual information) and ARI (Adjusted Rand Index)~\cite{rand1971objective}, two commonly used measures for clustering evaluation.
ACC is used as the primary metric for ranking systems for both tasks.

\paragraph{Intent Classifier}
Evaluation of Task 2 requires training an intent classifier using the induced schema.
Following previous work demonstrating the effectiveness of training a simple classifier given fixed embeddings from a pre-trained sentence encoder in few-shot setting~\cite{zhang-etal-2021-effectiveness-pre, casanueva-etal-2020-efficient}, we train a logistic regression classifier on top of off-the-shelf, static \textsc{all-mpnet-base-v2} sentence embeddings from the \textsc{SentenceTransformers} library~\cite{reimers-gurevych-2019-sentence,song2020mpnet}.

\section{Data}

\begin{table*}[tbh]
\centering
\addtolength{\tabcolsep}{-3pt}
\scalebox{0.90}{
\begin{tabular}{lccccccc}
\toprule
Dataset & \# conv. & \# turns per conv. & \# words per turn & \# intents per domain \\
\midrule
MultiDoGO \cite{Peskov2019MultiDomainGD}  & 10,829 & 16.7 & 11.4 & 6.7 \\
\multiwoz{} \cite{budzianowski-etal-2018-multiwoz} & 10,437 & 13.7 & 15.4  & * \\
\sgd{} \cite{rastogi2020towards} & 22,825 & 20.3 & 11.7 & 2.3 \\
\midrule
 DSTC11-\oursinsuranceselfdstc{} & 948 & 70.5 & 12.3 & 22 \\
 DSTC11-\oursbankingdstc{} & 1,000 & 59.2 & 17.4 & 29 \\
 DSTC11-\oursfinancedstc{} & 2,000 & 65.1 & 15.7 & 39 \\
\bottomrule
\end{tabular}
}
\caption{Summary statistics of DSTC 11 Track 2 datasets and comparison with previous task-oriented dialogue datasets.
*User intents are not explicitly annotated in MultiWOZ, but are instead implicit to each domain.
}
\label{Tab:datasets}
\end{table*}

\label{sec:data}
Publicly available dialogue datasets have primarily focused on development and evaluation of task-oriented dialogue systems where conversations are representative of written human-to-bot (H2B) conversations adhering to restricted domains and schemas~\cite{budzianowski-etal-2018-multiwoz,rastogi2020towards}.
Such datasets are not designed to be reflective of the characteristics of human-to-human (H2H) conversations, and thus are unlikely to serve as a realistic test bed for evaluating systems designed to learn from natural conversations.
However, realistic live conversations are difficult to simulate due to the training required to convincingly play the role of an expert customer support agent in non-trival domains~\cite{chen-etal-2021-action} and the additional costs associated with collecting and annotating free-form synchronous conversations.

To address this gap, we introduce a benchmark dataset designed to emulate natural call center conversations between customers and customer support agents.
Each conversation emulates a two-party spoken-form customer support scenario corresponding to a generated scenario based on a combination of intents, slots, and complex conversational phenomena to encourage diversity and naturalness.
To naturally collect a wide variety of intents, participants were encouraged to depart from the original intents with additional requests related to each scenario.
The process of annotating conversations with reference intents was decoupled from the collection of conversations in order to mimic the manual process of designing an intent schema based on conversations.
Annotators shared an open intent label set that was periodically reviewed throughout the process to merge duplicate intents.

We provide three domains as part of the challenge track: \textit{Insurance} (used as development data), \textit{Personal Banking} and \textit{Finance} (used as evaluation data).
For evaluating Task 2, each domain also includes a H2B-style balanced test set containing utterances labeled with intents from a reference intent schema.
These test sets are collected independently from the H2H-style conversations.
Conversations are also labeled with automatic \textit{InformIntent} dialogue act predictions indicating potentially relevant turns for use in Task 2, though these are non-exhaustive and include utterances that are not relevant.

\begin{figure}[tb]
\includegraphics[width=0.5\textwidth]{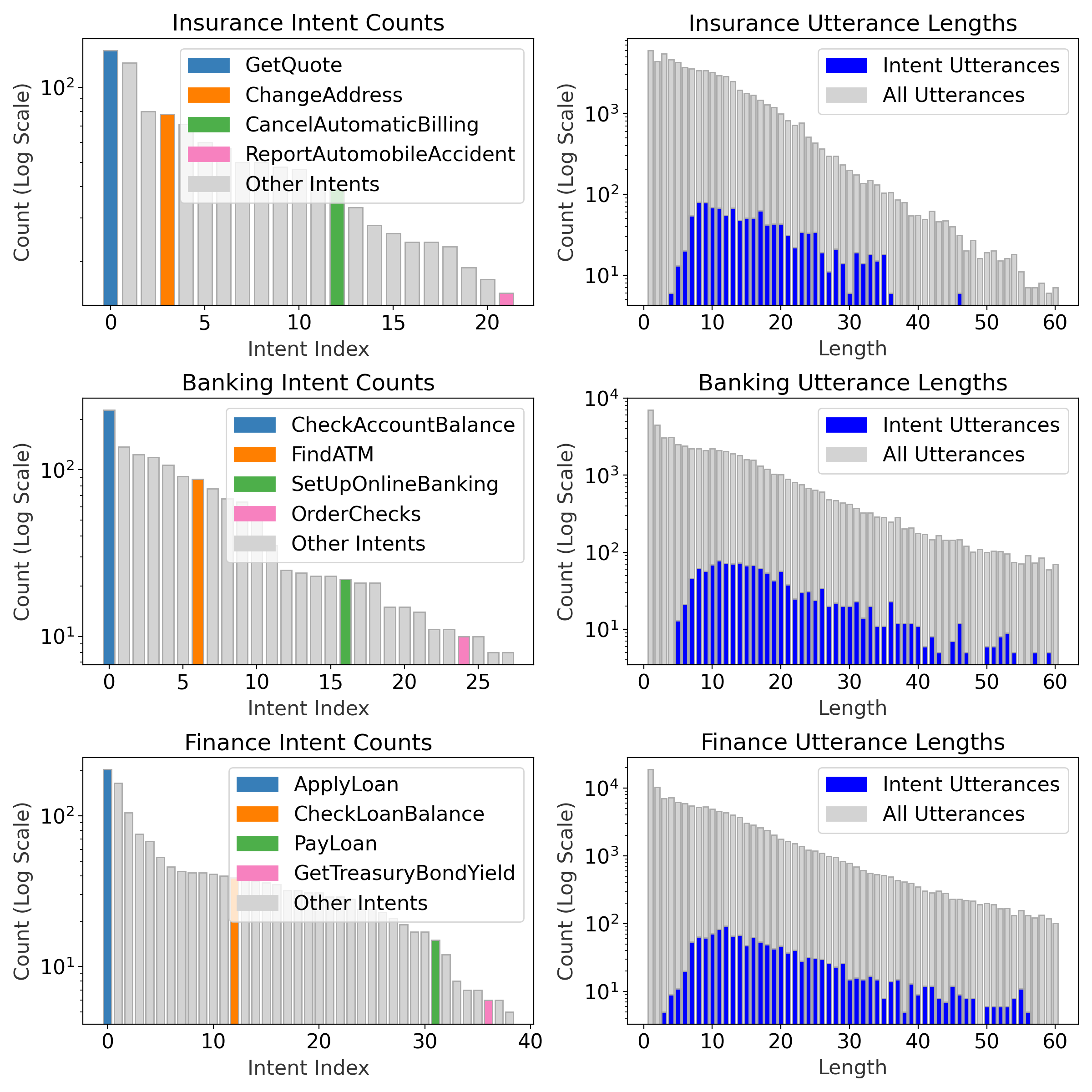}
\centering
\caption{Intent counts and utterance length distribution across domains (logarithmic scales).}
\label{Fig:intent-distribution}
\end{figure}

We present high-level statistics for each of the domains in comparison with pre-existing dialogue datasets in Table~\ref{Tab:datasets}. 
The resulting conversations include considerably more turns on average than previous task-oriented dialogue conversational datasets, suggesting a higher level of conversational complexity and diversity of flows, despite being restricted to single domains.
The conversations also contain a greater variety of intents per domain, ensuring the tasks are not trivial to distinguish between a small number of intents. 
Finally, as shown in Figure~\ref{Fig:intent-distribution}, the distribution of counts for these intents is highly skewed, similar to real intent distributions with long tails of infrequent intents and a few intents comprising a large volume of requests.
Fine-grained counts of intents for conversations and test sets and further analysis of intents are provided in Sections~\ref{sec:appendix:intent-counts} and \ref{sec:appendix:semantic-diversity}.
\section{Baselines}
\label{sec:baselines}

\paragraph{Task 1 Baseline}
The baseline system for Task 1 casts the problem as turn-level unsupervised clustering, adopting \textit{k}-means~\cite{macqueen1967some} as the clustering algorithm.
Utterances are encoded using a sentence embedding model from the \textsc{SentenceTransformers} library~\cite{reimers-gurevych-2019-sentence}, \textsc{all-mpnet-base-v2}, which fine-tunes MPNet~\cite{song2020mpnet} with a contrastive objective on a dataset consisting of 1 billion sentence pairs derived from a number of datasets.

Because the number of reference intents is not provided for the task, the baseline system identifies the number of intents automatically by selecting the value for $k$ that results in the highest intrinsic measure of clustering performance.
Silhouette values~\cite{rousseeuw1987silhouetttes,kaufman2009finding} indicate the appropriateness of a cluster assignment for a point based on the similarity to points in its assigned cluster and dissimilarity to points in other clusters.
An overall silhouette score can be computed by averaging the silhouettes for all points in a clustering.
The $k$ value that leads to the highest silhouette score is selected as the final result.
To accelerate the search for optimal clustering based on Silhouette scores, we employ sequential model-based global optimization following the tree-structured Parzen estimator (TPE) approach, implemented in the \textsc{Hyperopt} library~\cite{bergstra2013making}.

\paragraph{Task 2 Baseline}
The baseline system for Task 2 adopts the same clustering approach as Task 1.
Since clustering turns are not provided in this task, the baseline system uses the provided \InformIntent{} dialogue act classifier predictions to identify turns containing intents.

\section{Submissions}

\label{sec:Submissions}
We received submissions from 34 teams for Task 1 and 19 teams for Task 2 encompassing a wide range of techniques.
All teams that participated in Task 2 also participated in Task 1.
To preserve anonymity, teams are identified with IDs T0 through T37 (several team IDs were later removed after confirming duplicate submissions from different members of the same team).
This section provides a high-level summary of the submissions using self-reported descriptions and survey results of the participants.
A detailed overview of submissions is provided in appendix Table~\ref{tab:submissions-summary}.

\paragraph{Clustering Approach}
For both tasks, all submissions reported using clustering-based solutions.
Many submissions used clustering approaches that jointly learn input representations and cluster assignments. 
For example, at least ten submissions (T00, T02, T05, T07, T16, T17, T24, T30, T34, and T36) reported used SCCL~\cite{zhang-etal-2021-supporting}, a clustering approach that incorporates contrastive loss with Deep Embedded Clustering (DEC)~\cite{xie2016unsupervised}.
Other approaches (e.g. T06 and T37) separated representation learning from clustering, pre-training encoders with contrastive learning followed by K-Means clustering.
T06 used K-Means with representations from DSSCC~\cite{kumar-etal-2022-intent}.

Several submissions used HDBSCAN for clustering (T11, T13, T22, and T35), a density-based clustering algorithm that extends DBSCAN by allowing for clusters of varying density~\cite{campello2013density}.
Like DBSCAN~\cite{ester1996density}, it automatically categorizes outliers in low-density regions as noise, an appealing property for a task in which many noisy, outlying input utterances are expected to be present, though only one of these teams participated in Task 2 (T13).
Two teams (T13 and 35) performed dimensionality reduction on pre-trained embeddings using UMAP~\cite{mcinnes2018umap} prior to clustering, following BERTopic~\cite{grootendorst2022bertopic}.
Three other teams (T22, T27, and T31) also reported using some form of dimensionality reduction as part of their submissions.

\paragraph{Pre-trained Models}
Almost all submissions reported using specific pre-trained models, many of which were encoders tailored for computing utterance embeddings such as \textsc{all-mpnet-base-v2}~\cite{reimers-gurevych-2019-sentence}, \textsc{dse-bert-base}, \textsc{dse-roberta-large}~\cite{zhou-etal-2022-learning}, and \textsc{sup-simcse-roberta}~\cite{gao-etal-2021-simcse}.
Teams T00, T15, T23 and T34 used DSE-pretrained encoders~\citet{zhou-etal-2022-learning}, which learn sentence representations tailored for dialogues using contrastive learning with consecutive utterances in dialogues.
The majority of submissions also reported further fine-tuning encoders to tailor them to the task of intent induction (such as through SCCL, DEC, or supervised contrastive learning).
T11 reported using two pre-trained NLI models to pre-compute a distance matrix between inputs for HDBSCAN-based clustering, further finetuning one of these models on the task development data.

\paragraph{Use of External Data}
Roughly half of submissions reported using external public data for training their models.
These datasets included BANK77~\cite{casanueva-etal-2020-efficient}, CLINC150~\cite{larson-etal-2019-evaluation}, ATIS~\cite{hemphill-etal-1990-atis}, ACID~\cite{acharya2020using}, HWU64~\cite{liu2019benchmarking}, StackOverflow~\cite{predict-closed-questions-on-stack-overflow,xu-etal-2015-short}, SGD~\cite{rastogi2020towards} and MultiWOZ~\cite{budzianowski-etal-2018-multiwoz} and SNIPS~\cite{coucke2018snips}.
A number of teams reported using supervised pre-training (T06, T19, T23, T25, T30, T37) with labeled data, while others used semi-supervised approaches based on data augmentation to generate contrastive examples.
In addition to team T23 reported using automatic English machine translations of Chinese domain-specific insurance and financial data for pre-training.

\paragraph{Cluster Selection}
Because the number of intents was not provided, submissions using parameteric clustering approaches had to devise approaches for cluster selection.
Many submissions (12) explicitly reported using Silhouette scores.
Two submissions (T05 and T34) used HDBSCAN~\cite{campello2013density} to determine the number of clusters prior to using SCCL.
T15 reported using an iterative merging technique to determine the final number of clusters, perhaps similar to the strategy used in ~\citet{chatterjee-sengupta-2020-intent}.
T00 reported using a combination of the Elbow method~\cite{hardy1994examination} and Silhouette scores.

\paragraph{Use of Conversational Information}
Because complete conversations were provided as inputs to both tasks, participants had the option to make use of additional contextual signals for improving clustering. 
However, few teams reporting using conversational information beyond input turns (either given in Task 1, or predicted in Task 2) for clustering.
Some teams, such as T21 and T29, reported using input dialogues for continued pre-training of their models.
Team T28 identified the most relevant turn in the dialogue corresponding to each input turn using similarity scores from sentence embeddings, concatenating embeddings for the two turns as clustering inputs.

\paragraph{Task 2 Modifications}
The majority of teams that participated in Task 2 primarily used their Task 1 approaches, applying them to predicted \InformIntent{} turns instead of the provided clustering turns.
Although Task 2 provides \InformIntent{} predictions as suggested relevant turns, several teams reported developing their own models or approaches for identifying turns in Task 2.
Multiple teams (T02, T17, and T24) reported developing a custom classifier trained on public data to detect relevant turns or sentences containing intents.
Team T34 used a rule-based system to identify sentences for clustering, and further augmented the input data prior to applying SCCL for clustering.

\section{Results}
\label{sec:results}

In this section, we summarize the results for Task 1 and Task 2.
To aggregate results across datasets, we use simple averages.
As an alternative aggregate score, we compute the average mean reciprocal rank for ACC, \fone{}, and NMI, averaged over both datasets (\mrr{}).
This approach is intended to be less sensitive to scores of the individual datasets or biases of particular metrics.
Table~\ref{tab:ranking-summary} summarizes the aggregate results and team rankings for Task 1 and Task 2.
Detailed results for each dataset are provided in Section~\ref{sec:appendix:semantic-diversity}

\begin{table}[tb!]
\centering
\addtolength{\tabcolsep}{-3pt}
\scalebox{0.90}{
\begin{tabular}{lcccc}
\toprule
\multirow{2}*{\textbf{Team}} & \multicolumn{2}{c}{\textbf{Task 1}} & \multicolumn{2}{c}{\textbf{Task 2}} \\
     & ACC & \mrr{}  & ACC  & \mrr{} \\
\midrule
     T23 &     \textbf{69.8 (1)} &     38.2 (4) &     \textbf{76.3 (1)} &     56.2 (2) \\
     T07 &     69.6 (2) &     41.0 (2) & -            & -            \\
     T35 &     69.3 (3) &     39.2 (3) & -            & -            \\
     T05 &     69.1 (4) &     \textbf{58.9 (1)} &     74.5 (5) &     31.7 (5) \\
     T02 &     68.8 (5) &     32.5 (5) &     75.3 (2) &     33.9 (4) \\
     T17 &     67.1 (6) &    12.8 (10) &     73.8 (6) &     \textbf{61.0 (1)} \\
     T36 &     66.3 (7) &     13.1 (8) &     74.9 (3) &     25.6 (6) \\
     T24 &     66.2 (8) &     19.9 (6) &     74.7 (4) &     36.1 (3) \\
     T00 &     64.9 (9) &     13.1 (9) &    59.7 (17) &     9.8 (12) \\
     T34 &    63.7 (10) &    10.7 (11) &    59.3 (19) &     5.9 (18) \\
     T25 &    62.6 (11) &     9.3 (12) & -            & -            \\
     T10 &    62.5 (12) &     6.0 (16) & -            & -            \\
     T14 &    61.5 (13) &     6.0 (17) &     69.6 (7) &     11.8 (8) \\
     T30 &    61.1 (14) &     6.7 (14) &    60.9 (16) &     7.0 (16) \\
     T19 &    59.2 (15) &     5.9 (18) &    67.5 (10) &    10.1 (11) \\
     T37 &    58.7 (16) &     5.8 (19) & -            & -            \\
     T29 &    58.1 (17) &     5.6 (21) & -            & -            \\
     T26 &    58.0 (18) &     4.8 (24) &    63.6 (12) &     7.4 (14) \\
     T06 &    57.6 (19) &     4.4 (27) & -            & -            \\
     T01 &    57.4 (20) &     7.5 (13) & -            & -            \\
     T20 &    56.8 (21) &     5.6 (20) &    64.4 (11) &     10.9 (9) \\
     T28 &    56.6 (22) &     6.5 (15) & -            & -            \\
     T03 &    55.6 (23) &     4.6 (26) &    62.4 (15) &     7.7 (13) \\
     T13* &    55.3 (24) &     5.4 (22) &     69.4 (8) &    10.2 (10) \\ 
     T33 &    54.8 (25) &     14.7 (7) & -            & -            \\
     T18 &    54.8 (26) &     4.8 (23) & -            & -            \\
     T21 &    53.8 (27) &     3.9 (28) & -            & -            \\
     T16 &    52.9 (28) &     3.6 (29) & -            & -            \\
     T04 &    50.1 (29) &     4.8 (25) &    63.6 (12) &     7.4 (14) \\
     T15 &    48.9 (30) &     3.3 (32) &    63.4 (14) &     7.0 (17) \\
     T27 &    48.9 (31) &     3.5 (30) &     68.7 (9) &     12.6 (7) \\
     T22 &    46.7 (32) &     3.4 (31) & -            & -            \\
     T11* &    44.2 (33) &     3.2 (33) & -            & -            \\
     T31 &    35.1 (34) &     3.0 (34) &    59.5 (18) &     5.7 (19) \\
     \midrule
     Baseline &    55.8 (23) &     4.2 (28) &    63.6 (12) &     7.4 (14) \\
\bottomrule
\end{tabular}
}
\caption{
Summary of results for both tasks.
The ranking of the submission for each metric is given in parentheses.
ACC ($\uparrow$) is clustering accuracy and \mrr{} ($\uparrow$) is the average mean reciprocal rank across datasets. *Did not assign labels to all inputs (see Table~\ref{tab:label-propagation}).
}
\label{tab:ranking-summary}
\end{table}

\subsection{Task 1 Results}
Team T23 was the overall winner for Task 1, with an honorable mention to Team T05 for having the highest \mrr{}.
\begin{itemize}
    \item Team T23 had the highest overall ACC, the highest ACC on Finance, and the highest Precision (P) on Banking, and the fourth highest \mrr{}.
    \item Team T07 had the highest overall \fone{} score, the 2nd highest overall ACC, and the highest \fone{} on Finance.
    \item Team T05 had the top \mrr{}, the highest NMI and ARI and fourth highest overall ACC, with the highest ACC, \fone{}, NMI and ARI on Banking.
\end{itemize}

\subsection{Task 2 Results}
Team T23 was also the overall winner for Task 2, with an honorable mention for T17 for having the highest \mrr{}.
\begin{itemize}
    \item Team T23 had the highest overall ACC, Recall and ARI, and the second highest \mrr{}, with the highest ACC, P, F1, NMI, and ARI on Banking.
    \item T02 had the second highest overall ACC, the second highest ACC on Banking and the 4th highest ACC on Finance.
    \item Team T17 had the top \mrr{}, the highest overall P, F1, and NMI, the highest ACC, P, F1, NMI and ARI on Finance, and the 6th highest overall ACC.
\end{itemize}

\subsection{Analysis}

\paragraph{Approaches for Top Teams}
SCCL~\cite{zhang-etal-2021-supporting}, or more generally, deep embedded clustering (DEC)~\cite{xie2016unsupervised}, were highly popular approaches among the top submissions for both tasks.
In fact, only one of the top ten submissions, T35 (ranked third overall by both ACC and \mrr{}), instead used HDBSCAN for clustering.
Silhouette scores were also a popular choice for cluster selection among the top-performing submissions, though T05 (top overall \mrr{} for Task 1) and T34 were notable exceptions that used HDBSCAN to identify the number of clusters prior to using SCCL.

Among the top-performing teams, several used \textsc{dse-roberta-large}~\cite{zhou-etal-2022-learning}, while many others used \textsc{all-mpnet-base-v2}~\cite{reimers-gurevych-2019-sentence} as the base encoder models.
Many top systems also used a variety of public datasets to further adapt their models prior to clustering, either through supervised pre-training (multi-task with intent classification loss, or supervised contrastive learning~\cite{gao-etal-2021-simcse}), self-supervised pre-training (e.g. via masked-language modeling~\cite{devlin-etal-2019-bert} or contrastive learning~\cite{zhou-etal-2022-learning}), or a combination of approaches.
Among the top teams using \textsc{dse-roberta-large}, compared to T00 and T34, team T23 (the overall top submission by ACC) reported further supervised and self-supervised pre-training on multiple public datasets.

\paragraph{Label Propagation for Noise Cluster (Task 1)}
\begin{table}[tb]
\centering
\scalebox{0.90}{
\begin{tabular}{cccc}
\toprule
Team &  LP &       ACC (Rank) &      \mrr{} (Rank) \\
\midrule
 \multirow{2}*{T11} &   & 44.2 (33) & 3.2 (33) \\
  & \checkmark &  65.8 (9) & 16.9 (7) \\
 \multirow{2}*{T13} &   & 55.3 (24) & 5.4 (22) \\
  & \checkmark & 57.2 (22) & 6.0 (16) \\
\bottomrule
\end{tabular}
}
\caption{
Impact of label propagation (LP) for on submissions that used HDBSCAN, but left noise instances unassigned.
For T11, the ranking changes from 33rd to 7th based on \mrr{}.
}
\label{tab:label-propagation}
\end{table}

We noticed that two submissions using HDBSCAN did not assign cluster labels to a number of instances in Task 1, instead leaving a large portion unassigned as automatically-detected noise (T11 and T13).
Task 1 evaluation penalizes this, since clustering evaluation assumes a label should be assigned to every input.
To validate this, we propagated non-noise cluster labels for these teams to the unlabeled instances  by training the classifier described for Task 2 on labeled instances, and applying the resulting model to the remaining noise instances.
As shown in Table~\ref{tab:label-propagation}, we observed improvements for both submissions, with a drastic improvement for team T11, with the ACC and \mrr{}-based rankings improving from 33rd to 9th and 7th respectively, indicating a potential shortcoming of clustering-based evaluation for comparing intent induction methods.

\paragraph{Sensitivity to Classifier (Task 2)}
\begin{table}[tb]
\centering
\addtolength{\tabcolsep}{-3pt}
\scalebox{0.90}{
\begin{tabular}{lcccc}
\toprule
\multirow{2}*{\textbf{Team}} & \multicolumn{2}{c}{\textbf{ACC}} & \multirow{2}*{\centering \textbf{Best Model}}\\
 &                               Avg. (Rank) &   Max (Rank) &           \\
\midrule
 T23 &  \textbf{\std{74.9}{1.5} (\std{1.0}{0.0})} &  \textbf{76.6 (1)} &    all-roberta-l \\
 T02 &  \std{73.7}{1.7} (\std{2.2}{0.4}) &  75.9 (2) & multiqa-mpnet-b \\
 T36 &  \std{72.3}{2.3} (\std{4.0}{0.9}) &  75.3 (3) &    all-roberta-l \\
 T24 &  \std{72.2}{1.8} (\std{4.3}{1.2}) &  74.7 (4) &     all-mpnet-b \\
 T05 &  \std{70.9}{2.5} (\std{5.6}{0.7}) &  74.5 (5) &     all-mpnet-b \\
 T17 &  \std{72.6}{1.4} (\std{3.9}{1.4}) &  73.9 (6) &    dse-roberta-b \\
 T14 &  \std{68.7}{1.7} (\std{7.1}{0.3}) &  71.7 (7) &    all-roberta-l \\
 T13 &  \std{66.3}{3.5} (\std{9.2}{1.8}) &  70.7 (8) &    all-roberta-l \\
 T27 &  \std{66.6}{2.7} (\std{9.0}{0.5}) &  70.5 (9) &    all-roberta-l \\
 T19 &  \std{66.4}{1.2} (\std{9.0}{1.0}) & 67.5 (10) &     all-mpnet-b \\
 T20 & \std{63.0}{1.4} (\std{12.6}{1.3}) & 64.8 (13) & multiqa-mpnet-b \\
 T15 & \std{58.0}{4.4} (\std{16.0}{1.2}) & 63.4 (15) &     all-mpnet-b \\
 T03 & \std{62.6}{0.6} (\std{13.1}{1.5}) & 63.4 (14) &  all-minilml12 \\
 T30 & \std{60.1}{1.0} (\std{15.4}{0.5}) & 61.4 (16) &    all-roberta-l \\
 T00 & \std{54.4}{4.7} (\std{18.0}{0.7}) & 60.6 (17) &    all-roberta-l \\
 T31 & \std{53.9}{5.1} (\std{18.6}{0.5}) & 60.2 (18) &    all-roberta-l \\
 T34 & \std{58.0}{1.3} (\std{16.9}{1.3}) & 59.3 (19) &     all-mpnet-b \\
\bottomrule
\end{tabular}
}
\caption{
\textbf{Task 2} average and maximum ACC with corresponding rankings (in parentheses) computed for nine classifier models using different utterance encoders.
Teams are ordered corresponding to their original rank based on ACC with \textsc{all-mpnet-base-v2}.
}
\label{tab:classifier-sensitivity}
\end{table}

Because Task 2 evaluation is dependent on the selection of a classifier, we also analyze the impact of the classifier on evaluation results.
To understand this impact, we compute Task 2 evaluation results for 9 different encoders (including SimCSE, DSE, and \textsc{sentence-transformers} variants, see Section~\ref{sec:appendix:classifiers}).
Table~\ref{tab:classifier-sensitivity} aggregates these results.
When using the best-performing (highest ACC) classifier for each system, we observe the rankings for the top ten systems do not change (though ranks 11-15 fluctuate).
Examining the average ACC across all classifiers, we observe that while the top system typically does not change, the standard deviation of the rankings for other systems falls between 0.3 and 1.8, indicating some degree of fluctuation.
This indicates that prediction-based evaluation may introduce some noise into the process, and using high-quality classifiers or a variety of models may make evaluation results more robust.

\section{Conclusions}
\label{sec:conclusions}
We presented task definitions, evaluation methods, datasets, and baselines for the DSTC 11 track on intent induction from conversations for task-oriented dialogue.
The track saw a variety of submissions from 34 teams, with 19 teams submitting entries for both tasks.
We summarized these track submissions and provided analysis of the trends and overall results.

The aim of the track is to provide a benchmark facilitating evaluation of methods for automatic induction of customer intents in the realistic setting of customer service interactions.
We hope that the benchmark and datasets will encourage new lines of research related to the analysis of human-to-human conversations.

\bibliography{anthology,custom}
\bibliographystyle{acl_natbib}

\appendix
\section{Appendix}
\label{sec:appendix}

\subsection{Intent Counts}
\label{sec:appendix:intent-counts}

To provide a benchmark that reflects a realistic setting of customer support conversations, we collected conversations from three domains (Insurance, Banking, and Finance), introduced in Section~\ref{sec:data}.
The Insurance domain was used solely as development data for the challenge, while Banking and Finance were used for evaluating systems.
In this section, we report further details on the characteristics of these datasets.

\begin{table}[tbh]
    \addtolength{\tabcolsep}{-3pt}
    \centering
    \begin{tabular}{lrr}
    \toprule
                     Intent &  Count &  Test Count \\
    \midrule
                    GetQuote &    150 &         181 \\
                EnrollInPlan &    132 &          32 \\
               ResetPassword &     82 &          29 \\
               ChangeAddress &     81 &          31 \\
                  CancelPlan &     74 &          29 \\
                  ChangePlan &     64 &          29 \\
                     PayBill &     63 &          34 \\
          ReportBillingIssue &     51 &          32 \\
                   FileClaim &     51 &         124 \\
                AddDependent &     51 &          33 \\
               CreateAccount &     48 &          30 \\
     RequestProofOfInsurance &     42 &          31 \\
      CancelAutomaticBilling &     40 &          31 \\
     UpdatePaymentPreference &     38 &          31 \\
         CheckAccountBalance &     34 &          29 \\
             RemoveDependent &     28 &          31 \\
      UpdateBillingFrequency &     24 &          29 \\
          CheckPaymentStatus &     24 &          31 \\
      ChangeSecurityQuestion &     24 &          29 \\
             GetPolicyNumber &     19 &          30 \\
                   FindAgent &     18 &          29 \\
    ReportAutomobileAccident &     15 &          28 \\
    \bottomrule
    \end{tabular}
    \caption{Insurance domain counts for 22 intents in conversations and Task 2 test data.}
    \label{tab:stats-insurance}
\end{table}
\begin{table}[tbh]
    \addtolength{\tabcolsep}{-3pt}
    \centering
    \begin{tabular}{lrr}
    \toprule
                     Intent &  Count &  Test Count \\
    \midrule
         CheckAccountBalance &    251 &          30 \\
       InternalFundsTransfer &    139 &          26 \\
        ExternalWireTransfer &    124 &          26 \\
                  FindBranch &    120 &          21 \\
               DisputeCharge &    108 &          32 \\
          OpenBankingAccount &     92 &          21 \\
                     FindATM &     92 &          26 \\
        ReportLostStolenCard &     77 &          21 \\
              GetBranchHours &     73 &          21 \\
            CloseBankAccount &     65 &          32 \\
         UpdateStreetAddress &     60 &          20 \\
                 UpdateEmail &     35 &          20 \\
     CheckTransactionHistory &     25 &          20 \\
        AskAboutTransferTime &     24 &          21 \\
           UpdatePhoneNumber &     23 &          20 \\
          SetUpOnlineBanking &     23 &          20 \\
                ReportNotice &     23 &           0 \\
               GetBranchInfo &     22 &           0 \\
          GetWithdrawalLimit &     21 &          20 \\
              RequestNewCard &     15 &           0 \\
        AskAboutCashDeposits &     15 &          10 \\
              GetAccountInfo &     14 &           0 \\
    CheckAccountInterestRate &     11 &           0 \\
        AskAboutTransferFees &     11 &           0 \\
                 OrderChecks &     10 &           0 \\
         AskAboutCardArrival &     10 &           0 \\
              OpenCreditCard &      8 &           0 \\
             AskAboutATMFees &      8 &           0 \\
         AskAboutCreditScore &      4 &           0 \\
    \bottomrule
    \end{tabular}
    \caption{Banking domain counts for 29 intents in conversations and Task 2 test data.}
    \label{tab:stats-banking}
\end{table}
\begin{table}[tbh]
    \addtolength{\tabcolsep}{-3pt}
    \centering
    \begin{tabular}{lrr}
    \toprule
                     Intent &  Count &  Test Count \\
    \midrule
                  ApplyLoan &    222 &         106 \\
        CheckAccountBalance &    183 &          30 \\
                GetLoanInfo &    111 &          20 \\
               MakeTransfer &     77 &          42 \\
          OnlineBankingInfo &     73 &          20 \\
          GetCreditCardInfo &     55 &          20 \\
        ScheduleAppointment &     46 &          86 \\
          UpdatePhoneNumber &     43 &          21 \\
                OpenAccount &     43 &          37 \\
            GetExchangeRate &     43 &          23 \\
        UpdateStreetAddress &     41 &          20 \\
    ChangeStatementDelivery &     41 &          25 \\
           CheckLoanBalance &     40 &          22 \\
                UpdateEmail &     39 &          26 \\
            ApplyCreditCard &     39 &          75 \\
                  ChangePin &     36 &          20 \\
               RequestEmail &     35 &          10 \\
             SetAutoPayment &     33 &          62 \\
      MakeCreditCardPayment &     33 &          28 \\
                CancelCheck &     32 &          22 \\
         GetDebtIncomeRatio &     31 &          22 \\
           AddUserToAccount &     31 &          23 \\
      AskConsumerPriceIndex &     29 &          36 \\
               CloseAccount &     28 &          23 \\
             GetBranchHours &     25 &          20 \\
                  NetIncome &     24 &          20 \\
                 OrderCheck &     23 &          25 \\
          AskLiquidityRatio &     21 &          22 \\
                 FindBranch &     19 &          10 \\
             RequestNewCard &     17 &          10 \\
           GetBankStatement &     17 &          48 \\
                    PayLoan &     16 &          22 \\
            GetTransactions &     12 &           0 \\
          GetPaymentDueDate &      8 &           0 \\
              GetStockQuote &      7 &          20 \\
        GetInvestmentReport &      7 &          24 \\
       GetTreasuryBondYield &      6 &          23 \\
            GetCreditReport &      6 &          25 \\
             PurchaseStocks &      5 &          22 \\
    \bottomrule
    \end{tabular}
    \caption{Finance domain counts for 39 intents in conversations and Task 2 test data.}
    \label{tab:stats-finance}
\end{table}

Tables~\ref{tab:stats-insurance}, \ref{tab:stats-banking}, and \ref{tab:stats-finance} show the counts of each intent appearing in conversations and Task 2 test data from Insurance, Banking, and Finance respectively.
Intents with fewer than 4 corresponding annotated utterances were excluded from each dataset.
Comparing intents in Banking and Finance domains, while there is overlap between the domains, it is clear that Banking focuses more on personal banking requests (such as checking account balances, transferring funds, or disputing charges.
In contrast, Finance focuses more on loans and investing.

As is evident from these counts, the distribution of intents is highly skewed, while the test set counts are more evenly distributed.
The decoupling of conversation annotations from test utterances provides a practical benefit for the external prediction-based evaluation of Task 2.
Collecting examples from a fixed set of intents as test data is easier than annotating input conversations directly.
In natural datasets, intents typically do not have balanced distributions.
Identifying long tail intents may be just as important (or more so) than identifying common intents.
Finally, annotating many conversations may still result in only a few examples of low-frequency intents, whereas collecting them directly ensures they are properly represented in the test data.

\subsection{Semantic Diversity}
\label{sec:appendix:semantic-diversity}
\begin{table*}[tb!]
\centering
\begin{tabular}{lcccc}
\toprule
 Dataset & Utts./Intent & Words/Utt. & Intents/Domain & Avg. Diversity \\
\midrule
\clincoos{}~\cite{larson-etal-2019-evaluation} & 157.0 & 8.5 & 15 & 27.4 \\
\polyaibank{}~\cite{casanueva-etal-2020-efficient}  & 169.9 & 13.4 & 77 & 23.2 \\
\midrule
\oursinsuranceselfdstc{} &  52.4 & 20.1 & 22 & 33.3 \\
\oursbankingdstc{} & 51.8 & 26.9 & 29 & 30.1 \\
\oursfinancedstc{} & 40.9 & 31.3 & 39 & 32.4 \\
\bottomrule
\end{tabular}
\caption{Comparison of intent utterances between track datasets and public intent classification datasets.
Avg. Diversity corresponds to semantic diversity described in Section~\ref{sec:appendix:semantic-diversity}.}
\label{Tab:intent-datasets}
\end{table*}
\begin{table*}[tb!]
\centering
\begin{tabular}{lrrrrr}
\toprule
              Intent &  DSTC11 &  MultiDoGO &  CLINC150 &  BANK77 &  SGD \\
\midrule
        CheckBalance &    \textbf{31.9} &                 17.9 &      27.8 &         & 23.1 \\
        MakeTransfer &    \textbf{34.3} &                 24.2 &      29.5 &         & 25.9 \\
ReportLostStolenCard &    \textbf{29.0} &                 18.6 &      16.2 &    18.4 &      \\
       DisputeCharge &    \textbf{35.3} &                 23.7 &      26.1 &         &      \\
         OrderChecks &    \textbf{31.8} &                 21.5 &      19.0 &         &      \\
    CloseBankAccount &    \textbf{26.4} &                 17.6 &           &    20.1 &      \\
 UpdateStreetAddress &    \textbf{31.4} &                 17.5 &           &    28.6 &      \\
           ChangePin &    \textbf{27.4} &                      &      20.3 &    19.7 &      \\
\bottomrule
\end{tabular}
\caption{Comparing semantic diversity for aligned intents across MultiDoGO~\cite{peskov-etal-2019-multi}, CLINC150~\cite{larson-etal-2019-evaluation}, BANK77~\cite{casanueva-etal-2020-efficient}, and SGD~\cite{rastogi2020towards}.}
\label{Tab:semantic-diversity}
\end{table*}

To be useful as an intent induction benchmark, utterances for intents should have a high degree of variation rather than rigidly follow the same structure in every conversation.
To measure this, we investigate the semantic diversity of intent turns following~\citet{casanueva-etal-2022-nlu}.
To compute semantic diversity for a single intent, we (1) compute intent centroids as the average of embeddings for the turns labeled with the intent using the SentenceBERT~\cite{reimers-gurevych-2019-sentence} library with the pre-trained model~\textsc{all-mpnet-base-v2}, then (2) find the average cosine distance between each individual turn and the resulting centroid.
Finally, (3) overall semantic diversity scores are computed as a frequency-weighted average over intent-level scores.

The semantic diversity scores for each domain as compared to CLINC150~\cite{larson-etal-2019-evaluation} and BANK77~\cite{casanueva-etal-2020-efficient}, along with high level statistics are provided in Table~\ref{Tab:intent-datasets}.
We observe that the semantic diversity for Insurance, Banking and Finance is higher than that of CLINC150 and Bank77, indicating greater potential modeling challenges.
We also compare the semantic of diversity of \ours{} with other datasets for specific aligned intents across datasets in Table~\ref{Tab:semantic-diversity}.

\subsection{Submission Overview}
\label{sec:appendix:submission-overview}

\begin{table*}[tbh]
\centering
\addtolength{\tabcolsep}{-3pt}
\scalebox{0.90}{
\begin{tabular}{c>{\centering\arraybackslash}m{0.15\linewidth}ccc>{\centering\arraybackslash}m{0.15\linewidth}>{\centering\arraybackslash}m{0.15\linewidth}>{\centering\arraybackslash}m{0.15\linewidth}cc}
\toprule
 \multirow{2}*{\textbf{Team}} & \multirow{2}*{\textbf{Clustering}} & \multicolumn{3}{c}{\textbf{Techniques}} & \multirow{2}*{\textbf{Selection}} & \multirow{2}*{\textbf{Base Model(s)}} & \multirow{2}*{\textbf{Datasets}} & \multicolumn{2}{c}{\textbf{Rankings}} \\
 & & SupPT & SSPT & Ens. & & & & Task 1 & Task 2 \\
\midrule
 T23 & DEC & \checkmark & \checkmark & & sil. & dse-roberta-l & BK CC DI SO & 01 / 04 & 01 / 02 \\
 T07 & SCCL & & \checkmark & & sil. & \textbullet & OD & 02 / 02 & \textbullet \\
 T35 & hdbscan & & & & sil. & SBERT & \textbullet & 03 / 03 & \textbullet \\
 T05 & SCCL & & & & hdb4k & all-mpnet-b & AD AS BK SO & 04 / 01 & 05 / 05 \\
 T02 & SCCL & \checkmark & \checkmark & & \textbullet & all-mpnet-b & AS BK SO & 05 / 05 & 02 / 04 \\
 T17 & SCCL & & & & \textbullet & all-mpnet-b & AD BK & 06 / 10 & 06 / 01 \\
 T36 & SCCL & & & \checkmark & \textbullet & all-mpnet-b & AD BK CC & 07 / 08 & 03 / 06 \\
 T24 & SCCL & & & & \textbullet & all-mpnet-b & AD BK CC & 08 / 06 & 04 / 03 \\
 T00 & SCCL & & & & elbow sil. & dse-roberta-l & N/A & 09 / 09 & 17 / 12 \\
 T34 & DEC SCCL & & & & hdb4k & dse-roberta-l & \textbullet & 10 / 11 & 19 / 18 \\
 T25 & k-means & \checkmark & & & sil. & all-mpnet-b & BK CC SGD & 11 / 12 & \textbullet \\
 T10 & centroid & & & & \textbullet & \textbullet & \textbullet & 12 / 16 & \textbullet \\
 T14 & centroid & & & & sil. & roberta-b & OD N/A & 13 / 17 & 07 / 08 \\
 T30 & SCCL & \checkmark & & & sil. & all-mpnet-b & BK CC HU & 14 / 14 & 16 / 16 \\
 T19 & DEC k-means & \checkmark & & & sil. & all-mpnet-b & AS BK CC OD HU MD SS & 15 / 18 & 10 / 11 \\
 T37 & k-means & \checkmark & & & sil. & SBERT & MW & 16 / 19 & \textbullet \\
 T29 & centroid & \checkmark & \checkmark & & sil. & \textbullet & \textbullet & 17 / 21 & \textbullet \\
 T26 & centroid & & & & sil. & \textbullet & N/A & 18 / 24 & 12 / 14 \\
 T06 & DSSCC k-means & \checkmark & \checkmark & & sil. & \textbullet & BK CC DI & 19 / 27 & \textbullet \\
 T01 & k-means & \checkmark & \checkmark & & \textbullet & \textbullet & \textbullet & 20 / 13 & \textbullet \\
 T20 & \textbullet & & & & \textbullet & sentence-t5-l & N/A & 21 / 20 & 11 / 09 \\
 T28 & k-means & & & & \textbullet & SBERT & \textbullet & 22 / 15 & \textbullet \\
 T03 & centroid & & & & sil. & sup-simcse-roberta roberta-b & BK & 23 / 26 & 15 / 13 \\
Base & \textbullet & & & & \textbullet & \textbullet & \textbullet & 23 / 28 & 12 / 14 \\
 T13 & hdbscan & & & & \textbullet & all-minilm-l6 & N/A & 24 / 22 & 08 / 10 \\
 T33 & k-means & \checkmark & \checkmark & & \textbullet & \textbullet & \textbullet & 25 / 07 & \textbullet \\
 T18 & k-means & \checkmark & \checkmark & & \textbullet & \textbullet & \textbullet & 26 / 23 & \textbullet \\
 T21 & centroid & \checkmark & \checkmark & & \textbullet & all-mpnet-b & CC DI & 27 / 28 & \textbullet \\
 T16 & SCCL & & & & \textbullet & \textbullet & \textbullet & 28 / 29 & \textbullet \\
 T04 & centroid & & & \checkmark & \textbullet & multi-minilm all-mpnet-b & \textbullet & 29 / 25 & 12 / 14 \\
 T15 & centroid & \checkmark & & & merge & dse-bert-b bart-b & BK CC MW & 30 / 32 & 14 / 17 \\
 T27 & DP-means hdbscan & & & \checkmark & \textbullet & all-mpnet-b pp-multilingual-mpnet-b & N/A & 31 / 30 & 09 / 07 \\
 T22 & hdbscan k-means & & & & \textbullet & \textbullet & \textbullet & 32 / 31 & \textbullet \\
 T11 & hdbscan & \checkmark & & & \textbullet & \textbullet & \textbullet & 33 / 33 & \textbullet \\
 T31 & DEC UVWB & & & & sil. & \textbullet & N/A & 34 / 34 & 18 / 19 \\
\bottomrule
\end{tabular}
}
\caption{
Summary of submissions with corresponding rankings for Task 1 and Task 2 (ACC rank / \mrr{} rank).
System descriptions are self-reported, as there was no open-source requirement for this challenge.
See Table~\ref{Tab:abbreviations} for abbreviation definitions.
}
\label{tab:submissions-summary}
\end{table*}

\begin{table*}[tbh]
\centering
\addtolength{\tabcolsep}{-3pt}
\begin{tabular}{c>{\centering\arraybackslash}p{0.75\textwidth}}
\toprule
 Abbreviation & Definition \\
\midrule
 N/A & No Extra Data \\
 AD & ACID~\cite{acharya2020using} \\
 AS & ATIS~\cite{hemphill-etal-1990-atis} \\
 BK & BANK77~\cite{casanueva-etal-2020-efficient} \\
 CC & CLINC150~\cite{larson-etal-2019-evaluation} \\
 DI & Development Data (Insurance) \\
 HU & HWU64~\cite{liu2019benchmarking} \\
 MD & MCID~\cite{arora2020cross} \\
 MW & MultiWOZ~\cite{budzianowski-etal-2018-multiwoz} \\
 SG & SGD~\cite{rastogi2020towards} \\
 SS & SNIPS~\cite{coucke2018snips} \\
 SO & StackOverflow~\cite{predict-closed-questions-on-stack-overflow,xu-etal-2015-short} \\
 \midrule
 multi-minilm & TingChenChang/hpvqa-lcqmc-ocnli-cnsd-multi-MiniLM-v2 \\
 dse-bert-b & aws-ai/dse-bert-base~\cite{zhou-etal-2022-learning} \\
 dse-roberta-l & aws-ai/dse-roberta-large~\cite{zhou-etal-2022-learning} \\
 bart-b & facebook/bart-base~\cite{lewis-etal-2020-bart} \\
 sup-simcse-roberta & princeton-nlp/sup-simcse-roberta~\cite{gao-etal-2021-simcse} \\
 roberta-b & roberta-base~\cite{liu2019roberta} \\
 all-minilm-l6 & sentence-transformers/all-MiniLM-L6-v2~\cite{reimers-gurevych-2019-sentence} \\
 all-mpnet-b & sentence-transformers/all-mpnet-base-v2~\cite{reimers-gurevych-2019-sentence} \\
pp-multilingual-mpnet-b & sentence-transformers/paraphrase-multilingual-mpnet-base-v2~\cite{reimers-gurevych-2019-sentence} \\
 sentence-t5-l & sentence-transformers/sentence-t5-large~\cite{reimers-gurevych-2019-sentence} \\
 \midrule
 centroid & Centroid-based Clustering \\
 hdbscan & HDBSCAN~\cite{campello2013density} \\
 k-means & K-Means~\cite{macqueen1967some} \\
 sil. & Silhouette Scores~\cite{rousseeuw1987silhouetttes,kaufman2009finding} \\
 DEC & Deep Embedded Clustering~\cite{xie2016unsupervised} \\
 SCCL & Supporting Clustering with Contrastive Learning~\cite{zhang-etal-2021-supporting} \\
 DSSCC & Deep Semi-Supervised Contrastive Clustering~\cite{kumar-etal-2022-intent} \\
 \midrule
elbow & Elbow method~\cite{hardy1994examination} \\
 hdb4k & HDBSCAN for identifying K \\
 merge & Iterative Merging \\
\bottomrule
\end{tabular}
\caption{Abbreviations for datasets, models, and techniques used by submissions.
}
\label{Tab:abbreviations}
\end{table*}

Table~\ref{tab:submissions-summary} provides a detailed overview of the track submissions.
Note that because there was no open source requirement for this challenge, all details are self-reported through a survey given to participants.
Thus some information may be missing or inaccurate (such as due to misinterpretations of descriptions).

\subsection{Detailed Results}
\label{sec:appendix:detailed-results}

\begin{table*}[tbh]
\centering
\addtolength{\tabcolsep}{-3pt}
\scalebox{0.90}{
\begin{tabular}{lccccccccccccccccc}
\toprule
\multirow{2}*{\textbf{Team}} & \multicolumn{7}{l}{\textbf{Banking}} & \multicolumn{7}{l}{\textbf{Finance}} & \multicolumn{3}{l}{\textbf{Avg.}} \\
 & {ACC} & {P} & {R} & {F1} & {NMI} & {ARI} & {K} & {ACC} & {P} & {R} & {F1} & {NMI} & {ARI} & {K} & {ACC} & {F1} & {NMI} \\
\midrule
T23 & 71.5 & \textbf{81.2} & 75.7 & 78.4 & 77.3 & 62.9 & 29 & \textbf{68.1} & 70.9 & 76.5 & 73.6 & 72.8 & 55.5 & 32 & \textbf{69.8} & 76.0 & 75.1 \\
T07 & 72.1 & 72.2 & 84.6 & 77.9 & 72.5 & 64.5 & 14 & 67.1 & 71.8 & 78.7 & \textbf{75.1} & 74.5 & \textbf{55.8} & 41 & 69.6 & \textbf{76.5} & 73.5 \\
T35 & 73.3 & 79.2 & 76.4 & 77.8 & 73.5 & 63.8 & 53 & 65.4 & \textbf{77.6} & 70.3 & 73.8 & \textbf{75.1} & 53.5 & 75 & 69.3 & 75.8 & 74.3 \\
T05 & \textbf{75.2} & 78.8 & 82.5 & \textbf{80.6} & \textbf{78.5} & \textbf{70.7} & 26 & 62.9 & 67.8 & 74.1 & 70.8 & 72.6 & 52.1 & 27 & 69.1 & 75.7 & \textbf{75.5} \\
T02 & 75.0 & 78.6 & 82.2 & 80.4 & 78.0 & 70.2 & 26 & 62.6 & 67.5 & 73.9 & 70.6 & 72.3 & 51.6 & 27 & 68.8 & 75.5 & 75.1 \\
T17 & 72.0 & 77.0 & 78.3 & 77.7 & 75.7 & 70.1 & 25 & 62.3 & 63.9 & 78.6 & 70.5 & 70.7 & 51.6 & 25 & 67.1 & 74.1 & 73.2 \\
T36 & 72.5 & 76.6 & 77.9 & 77.3 & 76.1 & 66.7 & 28 & 60.2 & 67.1 & 71.0 & 69.0 & 71.2 & 49.4 & 28 & 66.3 & 73.1 & 73.6 \\
T24 & 69.4 & 73.6 & 81.6 & 77.4 & 76.6 & 64.6 & 28 & 63.0 & 68.8 & 73.1 & 70.8 & 72.9 & 51.7 & 28 & 66.2 & 74.1 & 74.7 \\
T00 & 66.7 & 74.8 & 75.5 & 75.1 & 71.9 & 51.2 & 17 & 63.2 & 68.1 & 74.1 & 70.9 & 70.3 & 49.5 & 25 & 64.9 & 73.0 & 71.1 \\
T34 & 67.5 & 78.7 & 73.7 & 76.1 & 76.1 & 57.4 & 22 & 60.0 & 63.3 & 76.0 & 69.1 & 69.9 & 48.2 & 22 & 63.7 & 72.6 & 73.0 \\
T25 & 68.2 & 68.2 & 89.7 & 77.5 & 74.1 & 66.3 & 11 & 56.9 & 57.1 & 83.5 & 67.8 & 70.3 & 54.4 & 18 & 62.6 & 72.7 & 72.2 \\
T10 & 68.4 & 70.5 & 75.7 & 73.0 & 69.6 & 61.1 & 20 & 56.7 & 66.4 & 62.3 & 64.3 & 67.4 & 42.9 & 35 & 62.5 & 68.6 & 68.5 \\
T14 & 67.0 & 69.1 & 77.6 & 73.1 & 72.1 & 66.7 & 21 & 56.0 & 59.7 & 68.0 & 63.6 & 62.6 & 45.0 & 25 & 61.5 & 68.4 & 67.4 \\
T30 & 69.6 & 69.6 & 82.4 & 75.5 & 67.1 & 59.3 & 12 & 52.6 & 56.3 & 76.1 & 64.7 & 65.1 & 45.3 & 19 & 61.1 & 70.1 & 66.1 \\
T19 & 68.7 & 68.7 & 82.3 & 74.9 & 70.6 & 60.6 & 12 & 49.7 & 49.7 & 81.3 & 61.7 & 64.3 & 46.7 & 14 & 59.2 & 68.3 & 67.4 \\
T37 & 65.1 & 73.9 & 69.4 & 71.6 & 69.6 & 54.9 & 20 & 52.3 & 75.1 & 54.2 & 63.0 & 71.0 & 34.7 & 49 & 58.7 & 67.3 & 70.3 \\
T29 & 60.9 & 74.5 & 67.5 & 70.9 & 71.6 & 52.0 & 23 & 55.4 & 66.8 & 64.4 & 65.6 & 67.5 & 41.3 & 29 & 58.1 & 68.2 & 69.6 \\
T26 & 65.1 & 73.8 & 73.3 & 73.6 & 69.7 & 52.0 & 17 & 50.8 & 54.4 & 67.1 & 60.1 & 60.3 & 43.7 & 45 & 58.0 & 66.8 & 65.0 \\
T06 & 64.4 & 66.4 & 79.1 & 72.2 & 68.7 & 58.3 & 32 & 50.8 & 54.4 & 67.1 & 60.1 & 60.3 & 43.7 & 45 & 57.6 & 66.1 & 64.5 \\
T01 & 57.7 & 74.4 & 59.3 & 66.0 & 70.0 & 50.3 & 30 & 57.1 & 72.4 & 65.2 & 68.7 & 72.1 & 43.1 & 36 & 57.4 & 67.3 & 71.0 \\
T20 & 62.8 & 62.8 & \textbf{90.1} & 74.0 & 72.0 & 61.6 & 10 & 50.8 & 72.4 & 52.8 & 61.1 & 68.8 & 33.7 & 50 & 56.8 & 67.6 & 70.4 \\
T28 & 55.9 & 55.9 & 72.6 & 63.2 & 59.8 & 37.0 & 12 & 57.2 & 73.9 & 60.3 & 66.4 & 71.6 & 39.1 & 44 & 56.6 & 64.8 & 65.7 \\
T03 & 57.4 & 57.4 & 88.4 & 69.6 & 67.5 & 55.8 & 9 & 53.9 & 53.9 & 78.3 & 63.8 & 64.1 & 50.3 & 17 & 55.6 & 66.7 & 65.8 \\
T13 & 53.8 & 72.5 & 58.0 & 64.5 & 64.0 & 35.0 & 30 & 56.8 & 71.5 & 63.7 & 67.4 & 68.9 & 37.2 & 46 & 55.3 & 65.9 & 66.4 \\
T33 & 74.5 & 78.5 & 79.7 & 79.1 & 75.2 & 68.2 & 80 & 35.1 & 38.6 & 84.9 & 53.0 & 47.9 & 15.0 & 88 & 54.8 & 66.1 & 61.6 \\
T18 & 58.5 & 58.5 & 89.0 & 70.6 & 69.2 & 56.3 & 9 & 51.0 & 56.4 & 73.3 & 63.7 & 67.7 & 42.4 & 19 & 54.8 & 67.1 & 68.4 \\
T21 & 59.9 & 62.3 & 70.0 & 65.9 & 60.5 & 46.8 & 14 & 47.7 & 64.9 & 49.9 & 56.4 & 63.4 & 29.2 & 46 & 53.8 & 61.2 & 62.0 \\
T16 & 57.2 & 57.2 & 77.0 & 65.6 & 56.1 & 45.4 & 9 & 48.7 & 61.9 & 57.8 & 59.8 & 60.7 & 32.7 & 29 & 52.9 & 62.7 & 58.4 \\
T04 & 65.7 & 72.9 & 76.4 & 74.6 & 71.9 & 57.4 & 16 & 34.6 & 35.8 & \textbf{90.9} & 51.4 & 49.1 & 11.6 & 45 & 50.1 & 63.0 & 60.5 \\
T15 & 51.1 & 62.7 & 58.8 & 60.7 & 60.3 & 48.1 & 36 & 46.8 & 54.6 & 56.2 & 55.4 & 59.0 & 38.3 & 39 & 48.9 & 58.0 & 59.7 \\
T27 & 51.0 & 58.4 & 67.5 & 62.6 & 59.8 & 37.9 & 18 & 46.8 & 50.8 & 73.3 & 60.1 & 61.9 & 37.3 & 19 & 48.9 & 61.3 & 60.8 \\
T22 & 52.0 & 65.5 & 59.3 & 62.3 & 59.4 & 37.4 & 27 & 41.5 & 60.1 & 53.2 & 56.4 & 62.0 & 27.3 & 46 & 46.7 & 59.3 & 60.7 \\
T11 & 42.2 & 50.6 & 58.9 & 54.4 & 49.8 & 7.9 & 37 & 46.1 & 54.3 & 62.2 & 58.0 & 57.1 & 11.9 & 47 & 44.2 & 56.2 & 53.4 \\
T31 & 34.3 & 34.3 & 63.2 & 44.4 & 31.6 & 20.1 & 6 & 35.9 & 54.8 & 38.1 & 45.0 & 52.3 & 20.5 & 46 & 35.1 & 44.7 & 42.0 \\
\midrule
Base & 59.7 & 60.7 & 72.0 & 65.9 & 60.3 & 46.1 & 12 & 51.8 & 69.3 & 54.0 & 60.7 & 65.7 & 33.6 & 46 & 55.8 & 63.3 & 63.0 \\
\bottomrule
\end{tabular}
}
\caption{
Per-dataset results across all metrics for \textbf{Task 1}.
\textit{Base} indicates the baseline system described in Section~\ref{sec:baselines}.
K indicates the number of induced intents.
The number of reference intents for Banking and Finance are 29 and 39 respectively.
Bold denotes the best results for each dataset.
}
\label{tab:task1summary}
\end{table*}

\begin{table*}[tbh]
\centering
\addtolength{\tabcolsep}{-3pt}
\scalebox{0.90}{
\begin{tabular}{lccccccccccccccccccc}
\toprule
\multirow{2}*{\textbf{Team}} & \multicolumn{8}{l}{\textbf{Banking}} & \multicolumn{8}{l}{\textbf{Finance}} & \multicolumn{3}{l}{\textbf{Avg.}} \\
 & {ACC} & {P} & {R} & {F1} & {NMI} & {ARI} & {K} & {U/I} & {ACC} & {P} & {R} & {F1} & {NMI} & {ARI} & {K} & {U/I} & {ACC} & {F1} & {NMI} \\
\midrule
T23 & \textbf{88.7} & \textbf{92.1} & 93.6 & \textbf{92.9} & \textbf{94.2} & \textbf{84.8} & 26 & 141.7 & 63.9 & 65.2 & 86.1 & 74.2 & 80.7 & 58.9 & 39 & 170.1 & \textbf{76.3} & 83.5 & 87.4 \\
T02 & 79.4 & 82.6 & 90.2 & 86.2 & 89.1 & 72.3 & 35 & 47.8 & 71.3 & 73.8 & 86.2 & 79.5 & 85.5 & 65.5 & 36 & 56.7 & 75.3 & 82.9 & 87.3 \\
T36 & 78.4 & 84.5 & 87.2 & 85.9 & 89.0 & 72.0 & 42 & 107.7 & 71.3 & 73.1 & \textbf{88.0} & 79.8 & 85.0 & 70.7 & 42 & 161.2 & 74.9 & 82.8 & 87.0 \\
T24 & 77.9 & 85.0 & 87.7 & 86.3 & 89.4 & 72.6 & 42 & 103.6 & 71.5 & 74.5 & 87.5 & 80.5 & 85.5 & 68.8 & 42 & 155.5 & 74.7 & 83.4 & 87.5 \\
T05 & 77.9 & 86.0 & 87.7 & 86.8 & 90.4 & 73.5 & 39 & 93.1 & 71.2 & 73.0 & 87.3 & 79.5 & 85.3 & 67.0 & 40 & 140.3 & 74.5 & 83.2 & 87.9 \\
T17 & 75.2 & 87.5 & 84.5 & 86.0 & 89.5 & 71.6 & 40 & 91.5 & \textbf{72.4} & \textbf{79.0} & 85.2 & \textbf{82.0} & \textbf{86.8} & \textbf{71.3} & 43 & 130.8 & 73.8 & \textbf{84.0} & \textbf{88.1} \\
T09 & 73.0 & 85.5 & 84.8 & 85.1 & 89.8 & 71.0 & 36 & 46.6 & 70.1 & 73.1 & 86.5 & 79.2 & 83.9 & 64.9 & 37 & 54.9 & 71.5 & 82.2 & 86.8 \\
T14 & 75.9 & 77.6 & 89.9 & 83.3 & 87.0 & 70.4 & 30 & 122.8 & 63.2 & 63.7 & 86.0 & 73.2 & 80.3 & 59.8 & 34 & 195.1 & 69.6 & 78.3 & 83.7 \\
T13 & 73.5 & 89.9 & 75.9 & 82.3 & 85.0 & 69.7 & 51 & 72.2 & 65.4 & 75.4 & 73.0 & 74.2 & 79.5 & 54.6 & 51 & 130.1 & 69.4 & 78.3 & 82.3 \\
T27 & 71.7 & 87.2 & 78.4 & 82.6 & 86.1 & 67.4 & 51 & 72.5 & 65.7 & 73.4 & 76.6 & 75.0 & 81.3 & 60.2 & 51 & 130.9 & 68.7 & 78.8 & 83.7 \\
T08 & 79.4 & 80.6 & 87.0 & 83.7 & 87.7 & 73.5 & 25 & 147.4 & 57.5 & 58.9 & 87.3 & 70.4 & 78.4 & 56.0 & 28 & 237.0 & 68.4 & 77.0 & 83.1 \\
T19 & 74.2 & 74.2 & 91.9 & 82.1 & 87.0 & 69.2 & 19 & 193.9 & 60.8 & 64.0 & 87.0 & 73.7 & 80.5 & 58.1 & 33 & 201.1 & 67.5 & 77.9 & 83.7 \\
T20 & 65.1 & 65.4 & \textbf{96.3} & 77.9 & 85.6 & 62.9 & 13 & 283.4 & 63.6 & 70.2 & 82.7 & 75.9 & 82.9 & 62.6 & 50 & 132.7 & 64.4 & 76.9 & 84.3 \\
T15 & 66.1 & 76.9 & 77.6 & 77.3 & 82.9 & 61.2 & 36 & 40.8 & 60.6 & 65.0 & 74.8 & 69.5 & 77.5 & 56.2 & 39 & 45.4 & 63.4 & 73.4 & 80.2 \\
T03 & 61.7 & 62.2 & 92.4 & 74.3 & 82.4 & 59.7 & 15 & 245.6 & 63.2 & 63.7 & 86.0 & 73.2 & 80.3 & 59.8 & 34 & 195.1 & 62.4 & 73.8 & 81.3 \\
T30 & 70.3 & 70.3 & 94.3 & 80.5 & 86.0 & 66.2 & 12 & 122.8 & 51.5 & 52.2 & 82.6 & 64.0 & 72.0 & 44.4 & 19 & 83.4 & 60.9 & 72.3 & 79.0 \\
T00 & 75.7 & 80.8 & 86.2 & 83.5 & 87.3 & 68.1 & 34 & 108.4 & 43.7 & 45.1 & 75.4 & 56.5 & 65.8 & 38.8 & 39 & 170.1 & 59.7 & 70.0 & 76.6 \\
T31 & 69.3 & 78.1 & 70.3 & 74.0 & 79.4 & 61.9 & 41 & 89.9 & 49.7 & 56.6 & 66.3 & 61.1 & 69.8 & 36.7 & 48 & 138.2 & 59.5 & 67.5 & 74.6 \\
T34 & 63.6 & 69.8 & 77.4 & 73.4 & 79.6 & 58.5 & 21 & 35.2 & 55.0 & 55.7 & 85.3 & 67.4 & 75.6 & 50.3 & 24 & 35.4 & 59.3 & 70.4 & 77.6 \\
\midrule
Base & 70.8 & 73.7 & 87.2 & 79.9 & 84.0 & 66.1 & 26 & 141.7 & 56.5 & 64.2 & 71.9 & 67.8 & 76.2 & 48.4 & 50 & 132.7 & 63.6 & 73.9 & 80.1 \\
\bottomrule
\end{tabular}
}
\caption{
Per-dataset results across all metrics for \textbf{Task 2}.
U/I indicates the number of utterances per intent.
K indicates the number of induced intents.
\textit{Base} gives the baseline system performance (equivalent to T04 and T26).
Bold denotes the best results for each dataset.
}
\label{tab:task2summary}
\end{table*}

Table ~\ref{tab:task1summary} provides a dataset-level detailed summary of results for Task 1 including the number of induced intents for each system. 
Table ~\ref{tab:task2summary} provides a dataset-level detailed summary of results for Task 2 including the number of induced intents and average number of sample utterances for each induced intent.

\subsection{Sentence Encoders used for Classifiers}
\label{sec:appendix:classifiers}
In this section, we enumerate the classifiers used in Section~\ref{sec:results} to examine sensitivity of Task 2 classifier sensitivity.
We used the following 9 sentence embeddings as static (not fine-tuned) features to logistic regression classifiers:
\begin{itemize}
\item sentence-transformers~\cite{reimers-gurevych-2019-sentence} \begin{itemize}
\item all-mpnet-base-v2
\item multi-qa-mpnet-base-cos-v1
\item all-roberta-large-v1
\item all-MiniLM-L12-v2
\end{itemize}
\item DSE~\cite{zhou-etal-2022-learning} \begin{itemize}
\item aws-ai/dse-roberta-large
\item aws-ai/dse-roberta-base
\item aws-ai/dse-bert-base
\end{itemize}
\item SimCSE~\cite{gao-etal-2021-simcse} \begin{itemize}
\item princeton-nlp/sup-simcse-roberta-large
\item princeton-nlp/sup-simcse-roberta-base
\end{itemize}

\end{itemize}

\end{document}